\documentclass[acmtog,authorversion,nonacm]{acmart}

\usepackage{booktabs} 
\usepackage{pifont}

\newcommand{\xmark}{\ding{55}}
\usepackage{graphicx}
\usepackage{amsmath}
\usepackage{hyperref}
\usepackage{booktabs} 
\usepackage{algpseudocode}
\usepackage{cleveref}
\usepackage{multirow}
\usepackage{subcaption}
\usepackage{enumitem}
\usepackage{xcolor}  
\usepackage{tcolorbox}  
\usepackage[ruled]{algorithm2e}

\definecolor{lightyellow}{RGB}{255,255,250}
\definecolor{darkyellow}{RGB}{255,255,170} 
\definecolor{darkblue}{RGB}{0,0,255}     
\definecolor{darkpurple}{RGB}{200,0,200}  
\definecolor{darkred}{RGB}{230,0,0}        
\definecolor{darkgreen}{RGB}{0,128,0}     

\acmJournal{TOG}

\begin{document}

\title{WorldCraft: Photo-Realistic 3D World Creation and Customization via LLM Agents}

\author{Xinhang Liu}
\email{xliufe@connect.ust.hk}
\affiliation{%
  \institution{The Hong Kong University of Science and Technology}
  \country{Hong Kong}
}

\author{Chi-Keung Tang}
\email{cktang@cse.ust.hk}
\affiliation{%
  \institution{The Hong Kong University of Science and Technology}
  \country{Hong Kong}
}

\author{Yu-Wing Tai}
\email{yu-wing.tai@dartmouth.edu}
\affiliation{%
  \institution{Dartmouth College}
  \country{USA}
}

\makeatletter
\let\@authorsaddresses\@empty
\makeatother

\renewcommand\shortauthors{Liu et al.}
\newcommand{\TODO}[1]{\textbf{\color{red}[TODO: #1]}}

\begin{abstract}
Constructing photorealistic virtual worlds has applications across various fields, but it often requires the extensive labor of highly trained professionals to operate conventional 3D modeling software. 
To democratize this process, we introduce \emph{WorldCraft}, a system where large language model (LLM) agents leverage procedural generation to create indoor and outdoor scenes populated with objects, allowing users to control individual object attributes and the scene layout using intuitive natural language commands.
In our framework, a coordinator agent manages the overall process and works with two specialized LLM agents to complete the scene creation:  \emph{ForgeIt}, which integrates an ever-growing manual through auto-verification to enable precise customization of individual objects, and \emph{ArrangeIt}, which formulates hierarchical optimization problems to achieve a layout that balances ergonomic and aesthetic considerations.
Additionally, our pipeline incorporates a trajectory control agent, allowing users to animate the scene and operate the camera through natural language interactions. Our system is also compatible with off-the-shelf deep 3D generators to enrich scene assets.
Through evaluations and comparisons with state-of-the-art methods, we demonstrate the versatility of WorldCraft, ranging from single-object customization to intricate, large-scale interior and exterior scene designs. This system empowers non-professionals to bring their creative visions to life.

\end{abstract}



\begin{teaserfigure}
\begin{center}
    \includegraphics[width=1.0\linewidth]{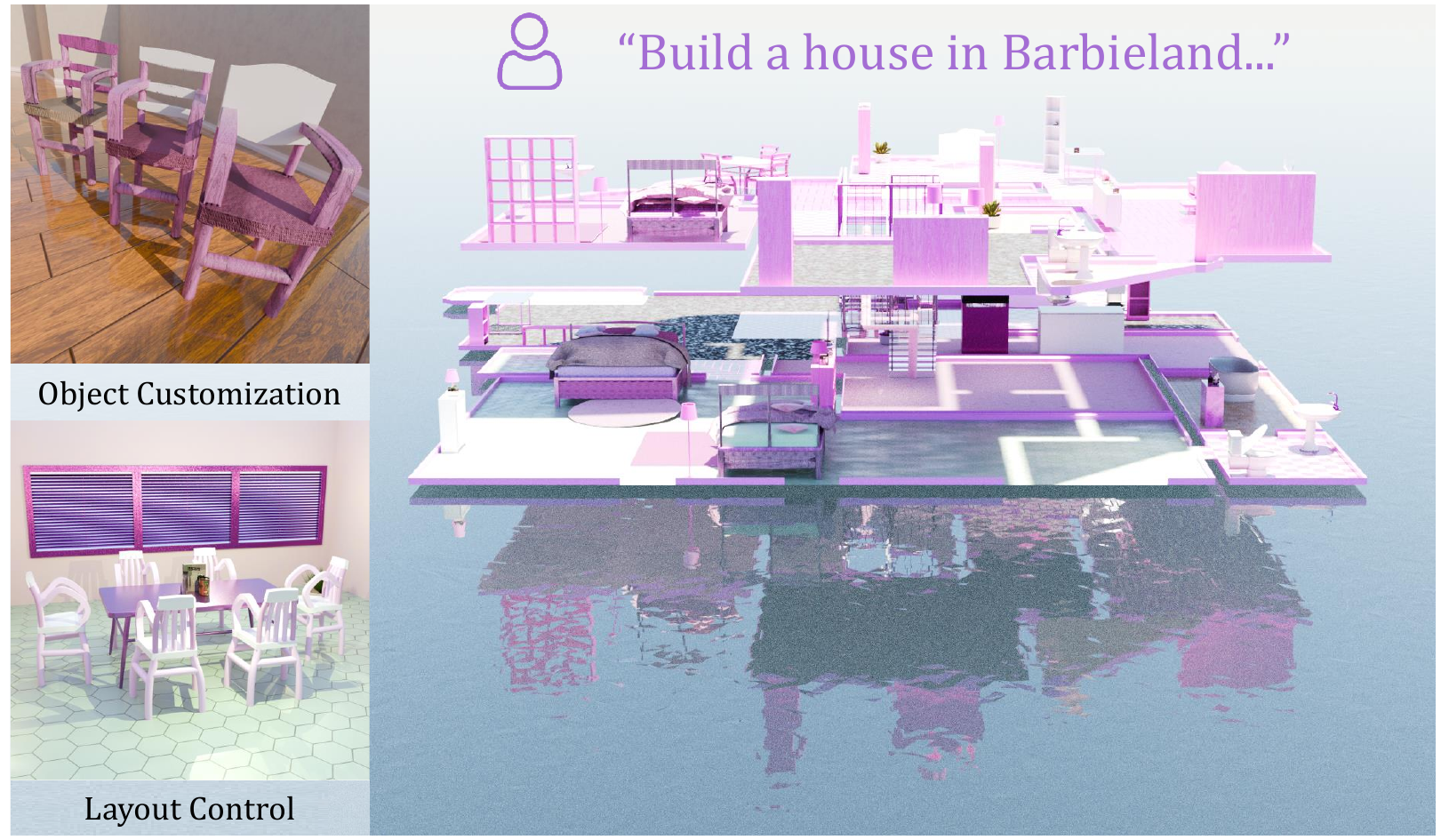}
\end{center}
\caption{
We introduce \emph{WorldCraft}, a system utilizing LLM agents to create complex, photo-realistic 3D virtual worlds from the user's text instructions. Our method populates these worlds with objects featuring precise geometry and PBR textures. Users can customize individual objects and 3D scene layouts through natural language interactions to achieve aesthetic, functional, and ergonomic designs.
}
\label{fig:teaser}
\end{teaserfigure}

\maketitle

\section{Introduction}
The creation of 3D virtual worlds offers extensive applications across entertainment and immersive technologies, including film, gaming, and mixed reality. It also facilitates advancements in robotics by providing simulated environments to train embodied agents~\cite{kolve2017ai2, srivastava2022behavior, xiang2020sapien,szot2021habitat, yang2024holodeck}.
However, constructing physically feasible, photorealistic virtual environments necessitates extensive human labor, requiring highly trained professionals to operate conventional 3D modeling software.

Recent approaches have begun to automate the process by generating 3D objects~\cite{poole2022dreamfusion,lin2023magic3d,hong2023lrm,zhang2024clay,siddiqui2024meta} or scenes~\cite{li2022infinitenature, yu2024wonderjourney, NEURIPS2023_7d62a85e, li2025dreamscene, zhou2025dreamscene360, po2024compositional} from text or single images. Nevertheless, state-of-the-art 3D scene generations often fall short in visual quality and lack detail. Furthermore, they typically do not offer granular customization options for individual objects or scene compositions with the desired layout.

To democratize the creation of photorealistic 3D worlds, we propose \emph{WorldCraft}, a system where LLM agents procedurally generate 3D scenes, emulating the step-by-step creative process of a human artist. As illustrated in \Cref{fig:teaser}, WorldCraft integrates users directly into the creation process, allowing them to use natural language to control individual objects and scene layouts, thereby assisting non-professionals in manifesting their creative visions.

Recent LLMs~\cite{achiam2023gpt, dubey2024llama, team2023gemini} have demonstrated remarkable capabilities, especially in managing complex visual tasks through programming~\cite{gupta2023visual}. However, they fall short in synthesizing 3D scenes, which requires a nuanced spatial understanding beyond simple token manipulation.
Procedural generators like Infinigen \cite{infinigen2023infinite, infinigen2024indoors} offer a glimpse of potential in LLM-based scene generation. Yet, due to lengthy pipelines and the complexity of parameter adjustments, off-the-shelf LLMs struggle to effectively manipulate these generators to tailor individual assets and arrange them according to the user’s design intent.

Recognizing these challenges, we propose a coordinator agent, which interacts with two specialized LLM agents to effectively navigate these complex procedural generators:
\textbf{(a) ForgeIt for individual object customization.} Procedural generators for individual objects within specific categories involve hundreds of parameters, which can appear nearly random to a general LLM agent. ForgeIt dynamically constructs a manual through an auto-verification mechanism. This ever-growing manual guides the agent in writing executable code to master the procedural generators, allowing it to accurately respond to user requests.
\textbf{(b) ArrangeIt for controllable scene layout generation.} To ensure that all objects in the scene are placed according to the user's design intent, as well as ergonomic factors such as visibility and accessibility, ArrangeIt formulates the scene arrangement as a hierarchical numerical optimization problem. This LLM agent then solves the problem using a novel optimization protocol.

Additionally, by incorporating a conversation-based trajectory control agent, our approach allows users to manipulate the movements of each object as well as the camera, thereby animating the world and synthesizing videos. Our 3D visual programming pipeline is also compatible with advanced off-the-shelf deep 3D generators ~\cite{hong2023lrm, instant3d2023, zhang2024clay, siddiqui2024meta}. These generators serve as complements to ForgeIt and introduce artistic objects that enhance the richness and diversity of the scenes.

Through a comprehensive evaluation and comparison with other state-of-the-art approaches, we demonstrate WorldCraft’s ability to interpret and execute complex 3D world creation instructions. Our user study further demonstrates its promising prospects in more practical applications.
In summary, our contributions include:
\begin{itemize}
    \item For the first time, we leverage LLM agents to procedurally generate highly complex and realistic indoor and outdoor 3D scenes.
    \item We introduce \emph{ForgeIt} for individual object control, constructing an ever-growing manual through an auto-verification mechanism.
    \item We propose \emph{ArrangeIt} for layout control, formulating scene arrangement as a hierarchical numerical optimization problem with a novel protocol to solve it.
    \item Our approach allows users to engage in intuitive natural language dialogues with the agent to customize individual objects, control layout, and direct movements.
\end{itemize}

\begin{figure*}[t]
\begin{center}
    \includegraphics[width=0.95\linewidth]{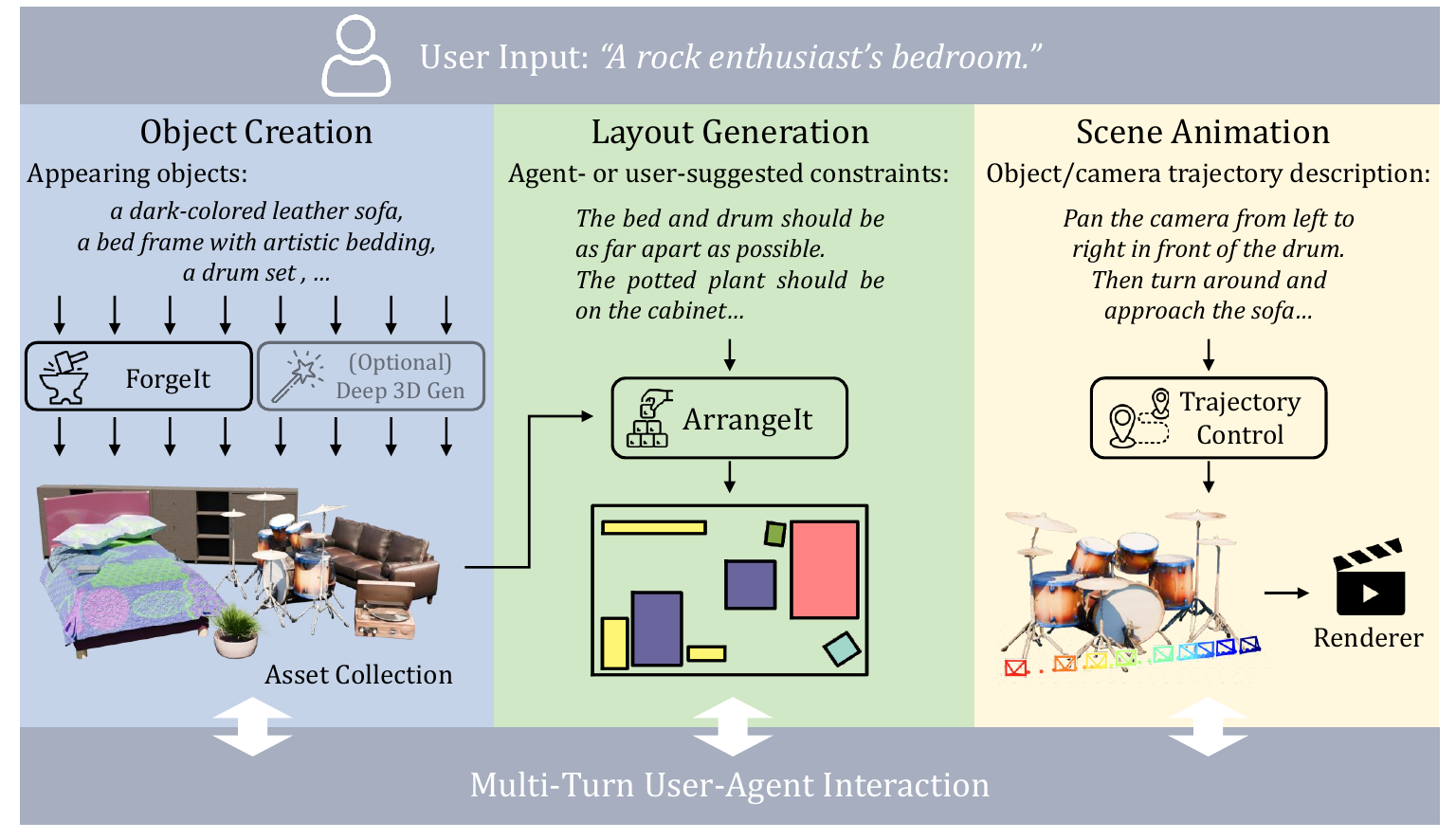}
\end{center}
\caption{\textbf{Overview of WorldCraft pipeline.} Starting with simple text input from the user, our coordinator agent creates a 3D scene in three stages: \emph{(a) Object creation.} The agent identifies objects that will appear in the scene and utilizes our ForgeIt system, or optionally, off-the-shelf deep 3D generators, to acquire the necessary assets.
\emph{(b) Layout generation.} 
The agent invokes our ArrangeIt module to design a layout that meets functional and aesthetic requirements.
\emph{(c) Scene animation.} Users can control objects or the camera trajectory through conversations to animate the scene and synthesize videos.
}
\label{fig:pipeline}
\end{figure*}

\section{Related Work}

\noindent\textbf{3D scene generation. }
Compared to the 3D generation of a single object, generating a complex 3D scene populated with multiple objects requires intricate detail modeling at various levels and a layout with both aesthetic and functional design considerations.
Earlier approaches \cite{devries2021unconstrained, bautista2022gaudi, chen2023scenedreamer, zhang2023berfscene} utilized generative models to capture the distribution of 3D scenes. Notably, \cite{liu2021infinite, li2022infinitenature} produced unbounded flythrough videos of natural scenes using GANs to render novel viewpoints, while \cite{hao2021gancraft} translated 3D semantic labels into radiance fields.
Recently, researchers have utilized 2D diffusion priors to synthesize 3D scenes. Specifically, studies such as \cite{NEURIPS2023_7d62a85e, yu2024wonderjourney, hoellein2023text2room, zhang2024text2nerf, li2025dreamscene} have employed an iterative process, using a 2D diffusion model to extrapolate scene content and lifting 2D images into 3D via depth estimation. Meanwhile, \cite{zhou2025dreamscene360} uses 2D diffusion models to create panoramic images from textual inputs, which are then transformed into 3D representations.
However, these approaches typically generate a single unified 3D representation of the entire scene, hindering object-level control and editability. 

In contrast, some works focus on compositional scene generation~\cite{zhai2023commonscenes, epsteindisentangled}.
\cite{Paschalidou2021NEURIPS} generates indoor scenes using an autoregressive transformer, \cite{po2024compositional} guides the generation of compositional 3D scenes based on user-specified layouts, and \cite{gao2024graphdreamer} employs a language model to create a scene graph for compositional 3D scene creation. 
\cite{yang2024holodeck} prompts an LLM to estimate spatial relations between objects to generate 3D environments for training embodied AI.
However, these methods often fail to handle object geometry and appearance adequately or rely on pre-existing 3D objects to compose the scene. Moreover, the synthesized object layouts usually do not accurately capture complex object relationships or respond precisely to user instructions. Thus, achieving functional, realistic 3D world creation with user-friendly customization remains an unresolved challenge.

\noindent\textbf{Layout generation.}
Layout generation is a critical step in compositional 3D scene creation, focusing on accurately estimating object coordinates and orientations to effectively arrange them within 2D or 3D spaces. Earlier approaches \cite{kjolaas2000automatic,coyne2001wordseye,germer2009procedural} utilized predefined templates and rules for this task. Notably, \cite{yu2011make} automatically generates indoor scene arrangements by extracting spatial relationships from user-provided exemplars. However, these methods heavily depend on extensive human input and struggle to generalize to new domains. 
Recent learning-based approaches \cite{wang2021sceneformer, Paschalidou2021NEURIPS, para2023cofs} achieve better robustness and generalizability using sequential modeling \cite{sun2025forest2seq} or denoising diffusion models \cite{tang2024diffuscene}.
Attempts have been made to leverage LLMs for indoor arrangements using textual descriptions \cite{fu2025anyhome, feng2024layoutgpt}. However, current approaches still struggle to interact effectively with users to fulfill their design intentions, often relying on demonstration exemplars for the agent to perform the task.
To enable 
user-friendly, complex 3D world creation, we propose a module that enables instruction-based layout generation, which allows users manipulation and control through easy natural language dialogues.

\noindent\textbf{AI agent and visual programming.}
Large Language Models (LLMs) have demonstrated remarkable capabilities in zero-shot and few-shot learning tasks across complex domains such as mathematics and commonsense reasoning~\cite{brown2020language, ouyang2022training, achiam2023gpt, touvron2023llama, dubey2024llama, team2023gemini}.
Some models are further enhanced by integrating visual capabilities, enabling them to handle tasks that combine linguistic and visual elements~\cite{alayrac2022flamingo, li2023blip, achiam2023gpt, liu2023llava}. Recent advancements have also shown that LLMs can interact with external tools to perform specific tasks~\cite{schick2023toolformer, wang2024internvid, shen2024hugginggpt, wangvoyager}. This ability extends to managing complex visual tasks by incorporating visual foundation models with LLMs,  or by translating visual queries into executable Python code~\cite{wu2023visual, gupta2023visual, suris2023vipergpt}, which proves promising in areas such as image generation and editing~\cite{sharma2024vision, lian2024llmgrounded, wu2024self, feng2024layoutgpt, wang2024genartist, yang2024mastering}. Particularly relevant to our study, SceneCraft~\cite{hu2024scenecraft} utilizes an LLM agent to translate text queries into 3D scenes by generating Blender scripts. However, it may fall short for higher scene complexity and visual quality necessary for more practical applications.

\section{Method}
To effectively turn a natural language user query into a detailed indoor or outdoor 3D scene that incorporates functional, ergonomic, and aesthetic considerations, WorldCraft employs a GPT-4 agent~\cite{achiam2023gpt} as the coordinator of the scene generation pipeline (\Cref{sec:coordinator}).

As illustrated in \Cref{fig:pipeline}, the generation pipeline consists of three primary stages: 
(a) \textbf{Object Creation}, where we identify objects to populate in the scene and utilize our proposed procedural asset generation agent, ForgeIt (\Cref{sec:forgeit}), or optionally, off-the-shelf deep 3D generators to gather the asset collection. 
(b) \textbf{Layout Generation}, in which the coordinator operates the ArrangeIt module (\Cref{sec:arrangeit}) to design a layout that satisfies both functional and aesthetic constraints based on the collected assets. 
(c) \textbf{Scene Animation}, where our trajectory control module enables users to guide the movements of objects or the camera through conversational inputs, allowing the scene to be animated and producing videos of comparable quality to those created by professional photographers (\Cref{sec:traj}).

\subsection{Scene Generation Coordinator}
\label{sec:coordinator}
The coordinator agent oversees the entire system. Its primary responsibilities involve decomposing complex tasks and solving them by invoking and collaborating with other agents. Additionally, it collects user feedback to meet their needs better.

\noindent\textbf{Task Decomposition.} Directly generating a complete 3D scene can be computationally expensive. Therefore, the coordinator agent breaks down the task into a series of manageable sub-tasks. For complex requests, such as generating an entire apartment with multiple rooms and numerous objects, instead of creating an overly complicated scene, the agent decomposes the environment into sub-spaces as a preprocessing step and then applies the same stages to each sub-space.

\noindent\textbf{Agent-Agent Interaction.} The coordinator agent works in collaboration with more specialized agents to complete the overall task. Specifically, in our system, there are agents dedicated to asset generation (ForgeIt), layout generation (ArrangeIt), and scene animation (Trajectory Control Agent). The coordinator assigns tasks to these agents and facilitates inter-agent communication to improve generation quality and overall coherence.

\noindent\textbf{User-Agent Interaction.} While the system primarily requires only a simple text prompt to begin, it supports extensive user engagement \emph{throughout all stages}. (Users can choose to disable intermediate interactions for a streamlined, end-to-end text-to-3D scene generation.) Users can customize individual objects, adjust layouts, and control movements through multi-turn, intuitive conversations, ensuring their design goals are met and refined throughout the iterative creative process.

In \Cref{fig:conversation}, we provide an example showcasing the user-agent and agent-agent interaction for creating a 3D scene.
\begin{figure}[t]
    \centering

{\small \begin{tcolorbox}[colframe=black, colback=lightyellow, coltitle=black, left=2mm, right=2mm]

\textcolor{darkblue}{\textbf{User}:} I want a recreation room in the university hall.

\textcolor{darkpurple}{\textbf{Coordinator}:} I will create a single room with a polished wooden floor, smooth white plaster walls, and a white plaster ceiling. Calling \texttt{create\_floor}, \texttt{create\_wall}, and \texttt{create\_ceiling}...

\textcolor{darkpurple}{\textbf{Coordinator}:} I will populate the scene with the following basic objects:
\begin{enumerate}[label=-, noitemsep, topsep=0pt]
    \item A comfortable black leather sofa.
    \item A metal desk lamp with an adjustable arm.
    \item [-] [More basic objects.]
\end{enumerate}

\textcolor{darkpurple}{\textbf{Coordinator}:} In addition, I will add the following advanced objects:
\begin{enumerate}[label=-, noitemsep, topsep=0pt]
    \item A green felt pool table with a wooden frame and brass corner pockets. 
    \item A vintage metal typewriter with exposed brass keys.
    \item [-] [More advanced objects.]
\end{enumerate}

\textcolor{darkpurple}{\textbf{Coordinator}:}  \colorbox{darkyellow}{\texttt{@User}} Is there anything else you would like to add?

\textcolor{darkblue}{\textbf{User}:} A simple wooden bookcase with five shelves, please.

\textcolor{darkpurple}{\textbf{Coordinator}:} Understood! I will now invoke the ForgeIt agent to generate the basic objects.

\textcolor{darkpurple}{\textbf{Coordinator}:} \colorbox{darkyellow}{\texttt{@ForgeIt}} Generate a comfortable black leather sofa.

\textcolor{darkred}{\textbf{ForgeIt}} \colorbox{darkyellow}{\texttt{@Coordinator}} Any specific requirements to make it comfortable?

\textcolor{darkpurple}{\textbf{Coordinator}:} \colorbox{darkyellow}{\texttt{@ForgeIt}} It should have a low back and thick cushions.

\textcolor{darkred}{\textbf{ForgeIt}:} Understood. Generating the asset...

[More interactions between the Coordinator and ForgeIt to generate other assets.]

\textcolor{darkpurple}{\textbf{Coordinator}:} Now I will use the deep 3D generator to create the advanced objects. Calling \texttt{deep\_3d\_gen}...

\textcolor{darkpurple}{\textbf{Coordinator}:} \colorbox{darkyellow}{\texttt{@ArrageIt}} Arrange these objects in the recreation room in the university hall: [The list of basic and advanced objects.]

\textcolor{darkgreen}{\textbf{ArrangeIt}:} Solving for the layout...

\textcolor{darkblue}{\textbf{User}:} \colorbox{darkyellow}{\texttt{@ArrangeIt}} Please do not place the felt pool table against the wall. It should be placed in the center of the space.

\textcolor{darkgreen}{\textbf{ArrangeIt}:}  Updating the layout...

\textcolor{darkpurple}{\textbf{Coordinator}:} Outputting the generated scene into ``scene.blend''.

\end{tcolorbox} }
 \caption{An example of user-agent and agent-agent interactions for decomposing tasks and collaboratively creating a 3D scene, demonstrating the system's capability to manage complex requests and facilitate user customization.}
    \label{fig:conversation}
    \end{figure}

\subsection{ForgeIt: Mastering 3D Object Generation through Auto-Verification}
\label{sec:forgeit}
Unlike prevalent deep learning-based text- or image-to-3D generation frameworks, ForgeIt utilizes an LLM agent to navigate procedural generators, specifically Infinigen~\cite{infinigen2023infinite, infinigen2024indoors}, to create a diverse array of 3D objects. It enables precise control and customization of object geometry and appearance through natural language interactions.

While LLMs excel in managing tools for specialized tasks, mastering complex procedural generators with numerous adjustable parameters is a challenge for general-purpose LLMs. To address this, ForgeIt constructs an ever-growing manual through an auto-verification mechanism. This manual guides the agent to iteratively master the use of procedural generation without the need for tedious human intervention or ground-truth labeling.

\begin{figure}[t]
\begin{center}
    \includegraphics[width=1.0\linewidth]{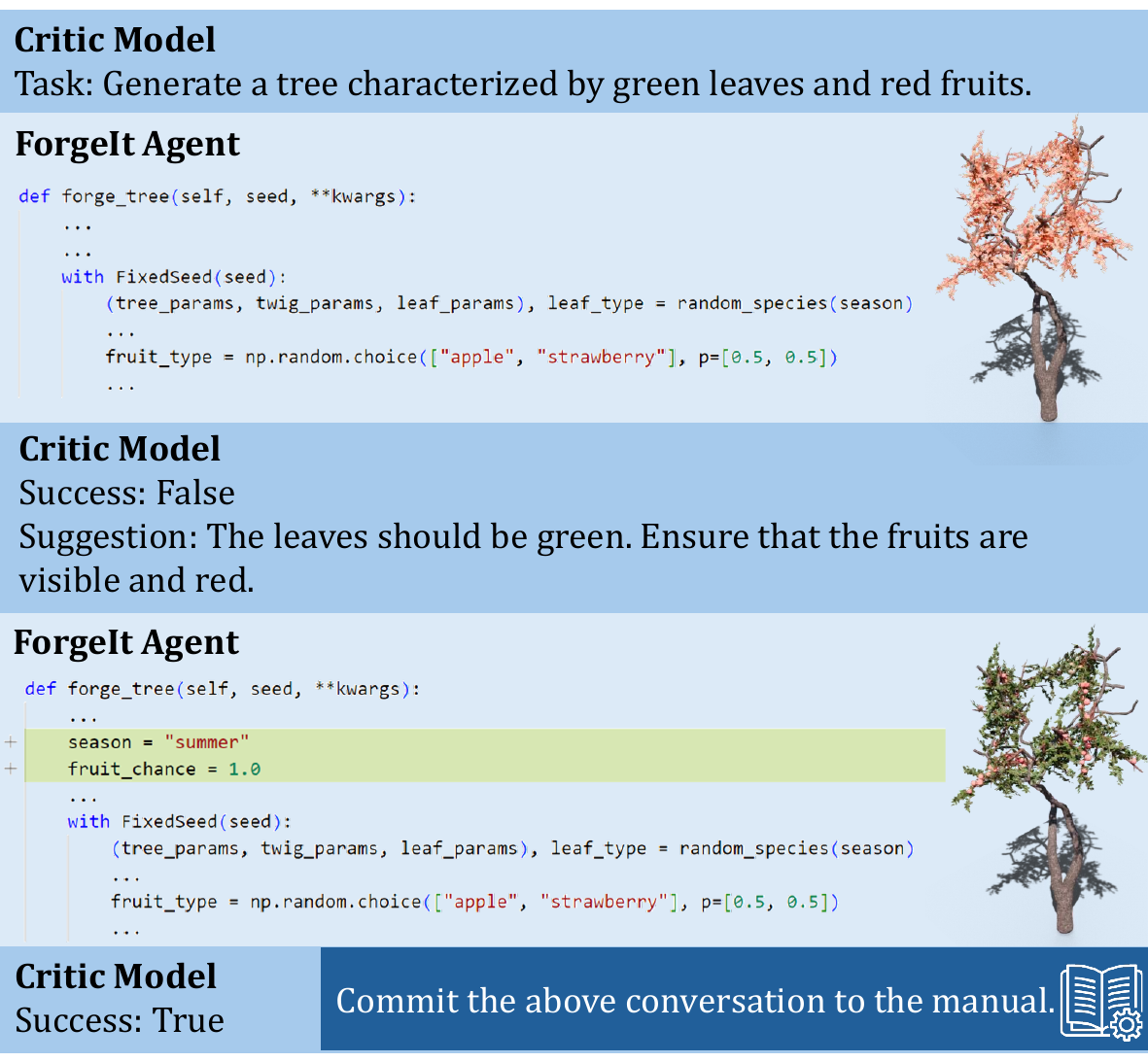}
\end{center}
\caption{\textbf{Manual construction procedure of ForgeIt.}
The critic model assigns the ForgeIt agent a text-to-3D generation task. The ForgeIt agent then synthesizes and executes a program in an attempt to generate the object. Subsequently, the critic model evaluates whether the generated object meets the task's requirements. If deemed successful, a record is committed to the manual.
}
\label{fig:forgeit_method}
\end{figure}

\noindent\textbf{Ever-growing manual.} We aim to dynamically construct a manual that the agent can reference when using a procedural generator for specific purposes. To achieve this, we employ another LLM as a critic model to facilitate auto-verification. 
Figure~\ref{fig:forgeit_method} illustrates the process where in each round, the critic model tasks the ForgeIt agent with generating an object, based on a textual description of the target. In response, the ForgeIt agent synthesizes and executes a program to meet these requirements. The resulting 3D object is then rendered from eight viewpoints. The critic model reviews the rendered images and determines whether the generated object meets the task requirements. If the task is evaluated as a failure, the critic model provides suggestions on how to improve the current program. This iterative process continues until the critic evaluates the outcome as successful, at which point the requirement and synthesized program are committed to the manual. If the agent fails to meet the requirements after a predetermined number of iterations, the system moves on to a new task.

Compared to the static method of exhaustively listing potential use cases and hard-coding them into prompts for an LLM, our dynamic approach of constructing an ever-growing manual more effectively coaches an agent to master procedural 3D object generation. This process utilizes an auto-verification mechanism with a critic LLM, thus eliminating the need for human intervention or explicit ground-truth labels as training signals.

\noindent\textbf{ForgeIt over deep 3D generators.} Unlike prevalent deep learning-based text- or image-to-3D generation frameworks~\cite{poole2022dreamfusion,lin2023magic3d,hong2023lrm,instant3d2023,zhang2024clay,siddiqui2024meta}, ForgeIt significantly reduces the computational demands associated with running large diffusion models or training on extensive 3D datasets.
Furthermore, ForgeIt avoids the post-processing step of mesh extraction using algorithms like Marching Tetrahedra~\cite{doi1991efficient}, which often results in poor surface geometry. Instead, ForgeIt directly crafts meshes through stringent mathematical rules, thereby enhancing compatibility with PBR pipelines. 
Additionally, by leveraging an LLM agent, ForgeIt inherently supports multi-turn conversational editing, allowing users to iteratively refine the generated asset to precisely meet specific requirements—an advantage not achieved by deep 3D generators.
While ForgeIt stands as a robust standalone tool, our framework remains compatible with these off-the-shelf deep 3D generators, whose optional involvement can enhance the creation by introducing more artistic objects.

\subsection{ArrangeIt: 3D Layout Control through Hierarchical Numerical Optimization}
\label{sec:arrangeit}
Given a collection of 3D assets, our goal is to arrange them while considering design objectives such as ergonomics, aesthetics, and functionality, and to allow users to control the arrangement through natural language instructions. To achieve this, we propose ArrangeIt, an approach where the agent models the scene arrangement as a set of hierarchical numerical optimization problems and solves them using a novel optimization protocol.

\begin{figure}[t]
\begin{center}
    \includegraphics[width=1.0\linewidth]{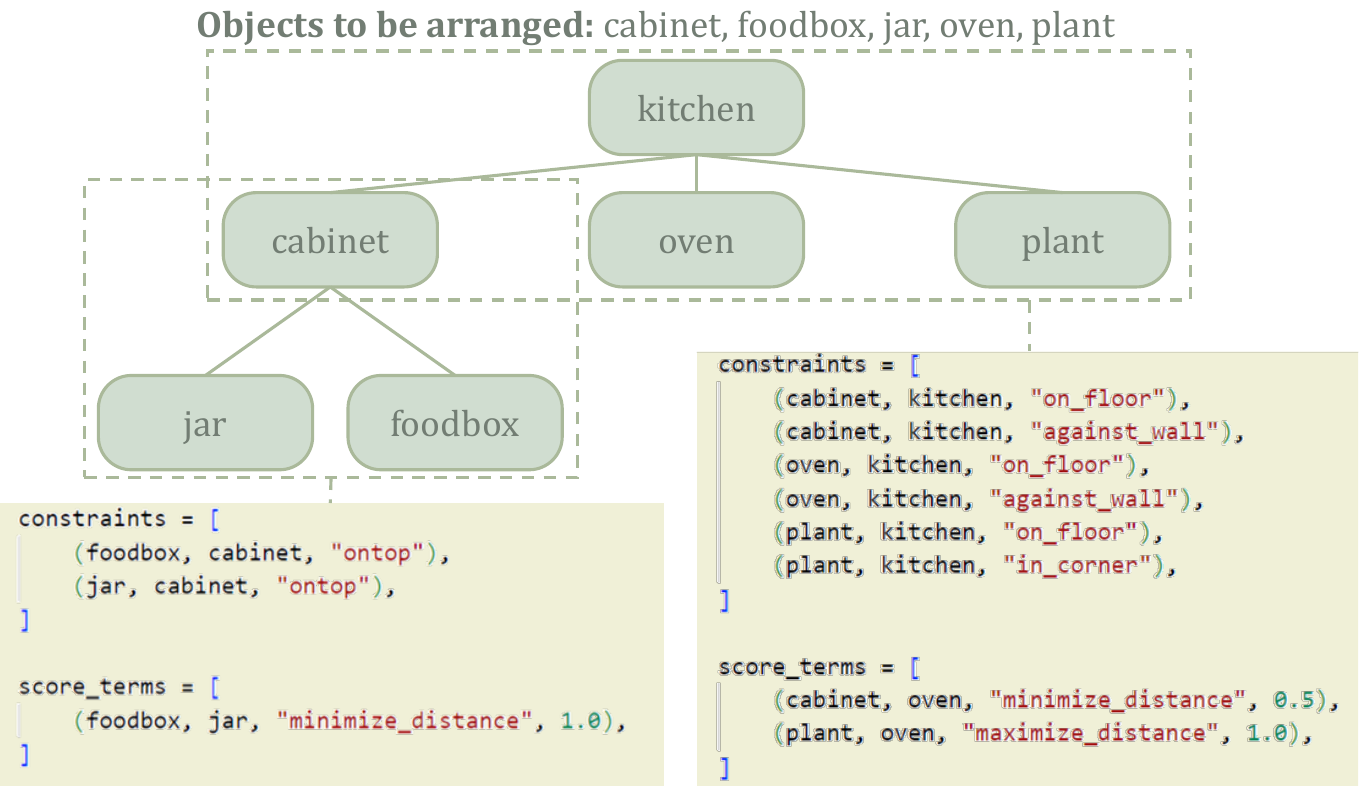}
\end{center}
\caption{\textbf{Formulation of the hierarchical numerical optimization  in ArrangeIt.} The agent constructs an object tree to hierarchically decompose the arrangement problem into subproblems, each of which is then modeled within our optimization protocol.} 
\label{fig:arrangeit_method}
\end{figure}

After the object creation stage, we typically have a large set of 3D assets, resulting in a prohibitive search space if we attempt to formulate a plan for all objects at once. Instead, we instruct the agent to recognize and leverage the hierarchical dependencies between objects—for example, a bookshelf and the books it holds. Specifically, the agent constructs an object tree and establishes subproblems to efficiently manage the complexity of the arrangement, as shown in Figure~\ref{fig:arrangeit_method}. 

Each of these subproblems is then formulated into a numerical optimization problem:
\begin{align*}
    \text{minimize} \quad & L(\{\mathbf{p}_i, \mathbf{\theta}_i\}_{i=1}^n) = \sum_{j=1}^{m} \lambda_j L_j(\{\mathbf{p}_i, \mathbf{\theta}_i\}_{i=1}^n) \\
    \text{subject to} \quad & c_1, c_2, \ldots, c_k,
\end{align*}
where the optimization variables are \(\{\mathbf{p}_i, \mathbf{\theta}_i\}_{i=1}^{n}\). \(\mathbf{p}_i = (x_i, y_i, z_i) \in \mathbb{R}^3\) represents the 3D location of object \(i\), and \(\mathbf{\theta}_i = (\theta_{ix}, \theta_{iy}, \theta_{iz}) \in [0, 2\pi]^3\) denotes its orientation in Euler angles.
The objective is a weighted sum of terms \(\{L_1, \ldots, L_m\}\), with their associated weights \(\{\lambda_1, \ldots, \lambda_m\}\). \(\{c_1, \ldots, c_k\}\) are the constraints that must be satisfied. The agent translates object relationships described in natural language into these objective terms and constraints to complete the formulation.

We have developed an optimization protocol that simplifies the coding of objectives and constraints for an LLM agent. Central to our approach is a series of API functions that reflect various spatial relationships and constraints, including:
\begin{itemize}
    \item \emph{Distance}: Measures the distance between two objects, based on their bounding boxes.
    \item \emph{Relative Orientation}: Calculates the difference in orientation between two objects.
    \item \emph{Alignment}: Aligns a set of objects along a specified axis.
    \item \emph{Proximity}: Ensures two objects are immediately adjacent to each other.
    \item \emph{Overlap}: Determines if two objects overlap along a specific axis.
    \item \emph{Symmetry}: Checks for rotational or reflection symmetry.
\end{itemize}
The protocol offers the flexibility to model spatial relationships, enabling their implementation as either hard constraints or soft constraints (score terms). For instance, regarding distance, a term can be incorporated into the objective to adjust the distance between objects, or hard constraints can be established to specify that the distance must be either greater than or less than a predefined value. This flexibility ensures that the LLM agent can effectively translate design intents into actionable layout directives.

After modeling the arrangement problem within our optimization protocol, we follow the methodology described in \cite{yu2011make} and employ simulated annealing~\cite{kirkpatrick1984optimization} with the Metropolis-Hastings criterion~\cite{hastings1970monte, metropolis1953equation} to find the optimal arrangement.

\subsection{Video Synthesis with Conversational Trajectory Control}
\label{sec:traj}
Upon completion of object creation and layout generation, the user can import the created world into software like Blender~\cite{blender} for rendering. While users can manually control the object trajectory by setting keyframes for their coordinates and orientations, our approach simplifies this process for users without experience in professional software, enabling them to direct the movements of objects and cameras using natural language.

We build our conversational trajectory control module upon ChatCam~\cite{liu2024chatcam}, a conversational camera control approach for NeRF and 3DGS representations. Our trajectory control module can be regarded as an extension of ChatCam in two key ways: (1) it supports mesh representation, and (2) it goes beyond just controlling the camera to include all objects in the scene.

Specifically, we follow ChatCam's methodology to extract scene-independent trajectory descriptions, and use an autoregressive text-to-trajectory model to translate them into trajectory commands. To place this trajectory within the scene, instead of ChatCam's image-based anchor determination procedure, we directly instruct the LLM to set anchors based on explicit object bounding boxes. 
For example, given a textual camera trajectory description like ``Pan the camera from left to right in front of the drum,'' our module first extracts the scene-independent part ``Pan the camera from left to right'' and translates it into a trajectory. The agent then uses bounding box information to correctly position this generated trajectory ``in front of the drum.''

\begin{figure*}[t]
\begin{center}
    \includegraphics[width=1.0\linewidth]{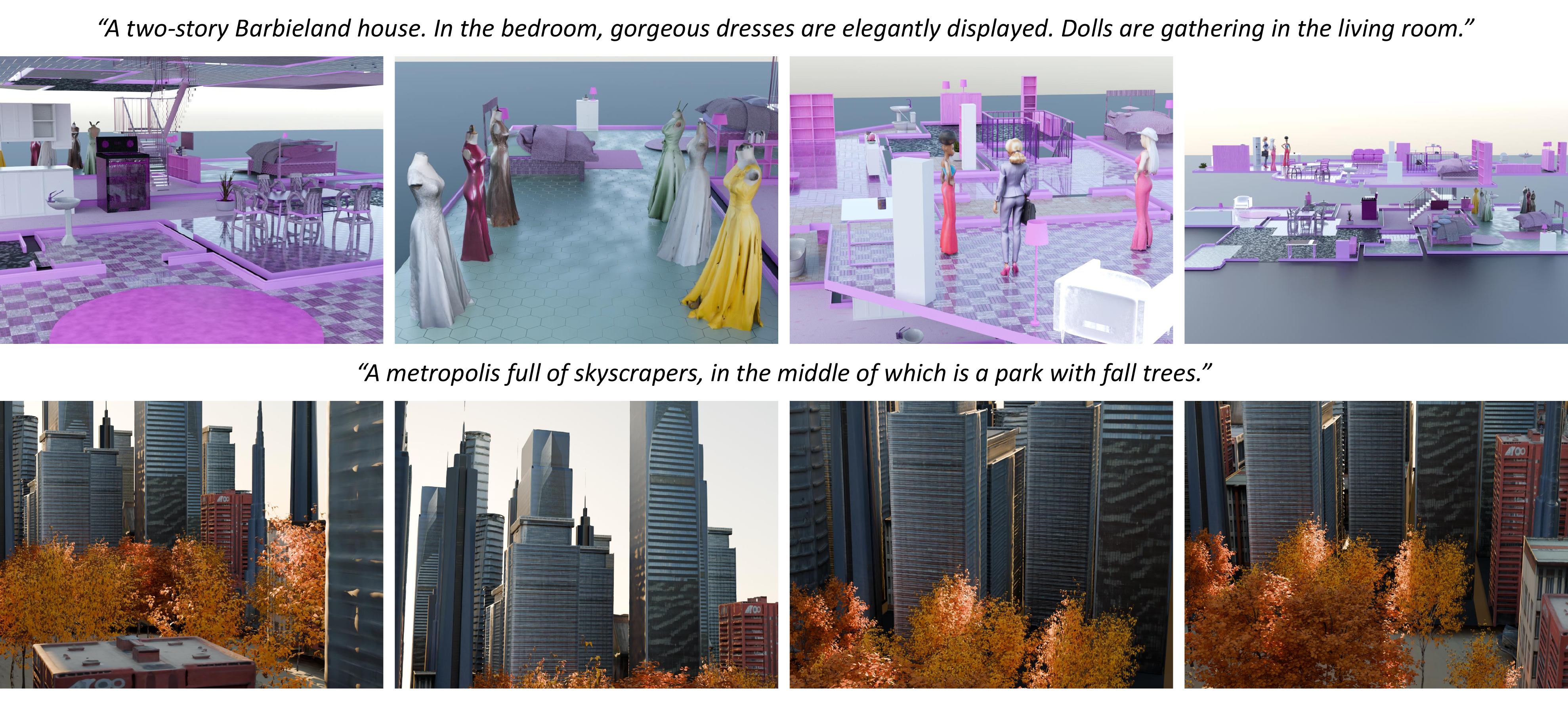}
\end{center}
\caption{\textbf{Language-guided complex scene generation.} Examples illustrating our method’s capability to generate expansive 3D indoor and outdoor scenes, richly populated with diverse objects.}
\label{fig:gallery_1}
\end{figure*}

\begin{figure*}[t]
\begin{center}
    \includegraphics[width=1.0\linewidth]{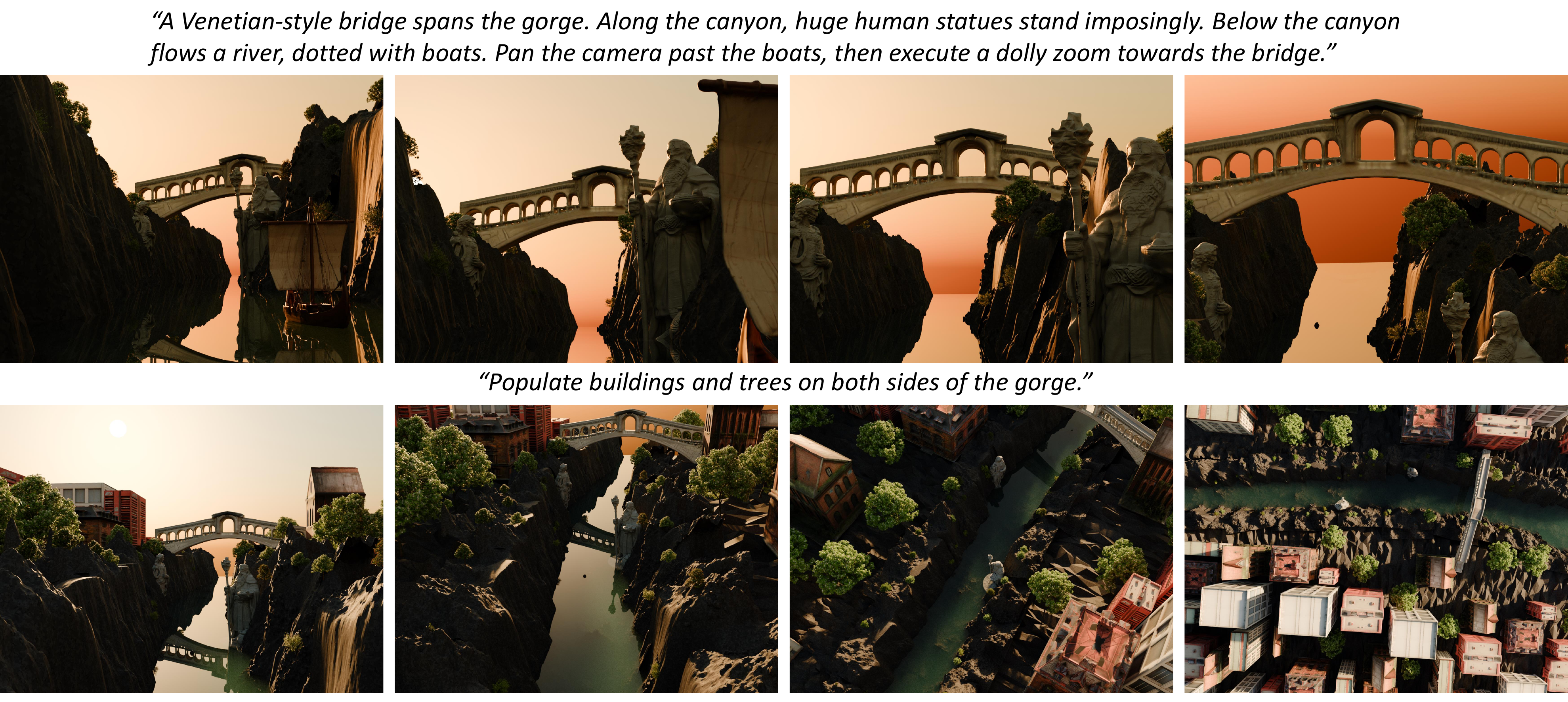}
\end{center}
\caption{\textbf{Language-guided scene generation and editing.}  Examples demonstrating our method's ability to generate complex large-scale outdoor scenes. Users can interact with our approach using natural language to further edit the created scenes.}
\label{fig:gallery_2}
\end{figure*}

\section{Experiments}
In this section, we evaluate the effectiveness of our proposed WorldCraft for 3D world creation across a range of challenging settings, comparing it qualitatively and quantitatively with state-of-the-art methods. Through evaluations and ablation studies, we provide empirical evidence of the effectiveness of its core modules. We then showcase our approach's capability to synthesize highly complex scenes.
We kindly refer the reader to our \emph{supplementary document and video} for additional experimental details and results.

\subsection{Experimental Setup}
\noindent\textbf{Implementation details.} We leverage OpenAI’s \textsc{gpt-4-0314}~\cite{achiam2023gpt} as both our agent and the critic model in the ForgeIt system. The ForgeIt agent navigates Infinigen~\cite{infinigen2023infinite,infinigen2024indoors} to procedurally generate 3D assets. We use Meshy\footnote{https://www.meshy.ai/api} as our additional deep 3D generator. CineGPT~\cite{liu2024chatcam}, originally designed for camera control, now serves as our text-to-trajectory model in the trajectory control module, is utilized without fine-tuning for general objects.

\noindent\textbf{Evaluation metrics.}
We evaluate the generated scene from three aspects: consistency with the input text, aesthetics (whether it is realistic and visually pleasing), and functionality (whether it respects ergonomics). Each of these aspects is rated on a scale from 1 to 10 by both users and the GPT-4 model. For consistency, we additionally report a CLIP score measuring the similarity between the rendered image of the generated scene and the input text. We also report the approach's runtime for synthesizing a single scene.

\subsection{Complex Scene Generation}
To better showcase the exceptional capabilities of our method in generating highly complex scenes, we present compelling examples in \Cref{fig:gallery_1} and \Cref{fig:gallery_2}.
In the first example in \Cref{fig:gallery_1}, our approach synthesizes a large, fully-furnished house tailored to the user-specified style. It further demonstrates our capability to adhere to user instructions by invoking a deep 3D generator to integrate additional objects into the scene seamlessly. The second example in \Cref{fig:gallery_1} depicts a cityscape where skyscrapers are neatly arranged beside a park filled with golden, autumnal trees.
\Cref{fig:gallery_2} showcases our method's proficiency in creating expansive outdoor scenes, featuring procedurally generated natural elements alongside artistically crafted objects from the deep 3D generator. This example also allows the user to edit further and enhance the scene using natural language, illustrating the adaptability of our interface.
This example also highlights our trajectory control module's capability of turning user instructions into corresponding object movements (see our supplementary video).

\begin{table}[t]
\centering
\caption{\textbf{Quantitative comparison on 3D scene generation.} Our approach achieves the best performance in terms of consistency, aesthetics, and functionality.}
\label{tab:comparison}
\resizebox{\linewidth}{!}{
\begin{tabular}{@{}lcccccccc@{}}
\toprule
\multirow{2}{*}{Method} & \multicolumn{3}{c}{Consistency ($\uparrow$)} & \multicolumn{2}{c}{Aesthetics ($\uparrow$)} & \multicolumn{2}{c}{Functionality ($\uparrow$)} & \multirow{2}{*}{Runtime ($\downarrow$)} \\
\cmidrule(r){2-4} \cmidrule(r){5-6} \cmidrule(r){7-8} 
 & CLIP & GPT-4 & User & GPT-4 & User & GPT-4 & User & \\
\midrule
\cite{yang2024holodeck} & 0.322 &  6.00 & 5.30 & 5.00 & 5.18 & 6.50 & 7.82&\textbf{5 min} \\
\cite{li2025dreamscene} & 0.281 &  7.00 & 5.11 & 6.00 & 6.72 & 7.00 & 4.31 & 53 min \\
Ours & \textbf{0.384} & \textbf{8.50} & \textbf{6.39} &  \textbf{8.00} & \textbf{7.15} & \textbf{7.00} & \textbf{8.01} & 18 min \\
\bottomrule
\end{tabular}
}
\end{table}

\begin{figure*}[t]
\begin{center}
    \includegraphics[width=1.0\linewidth]{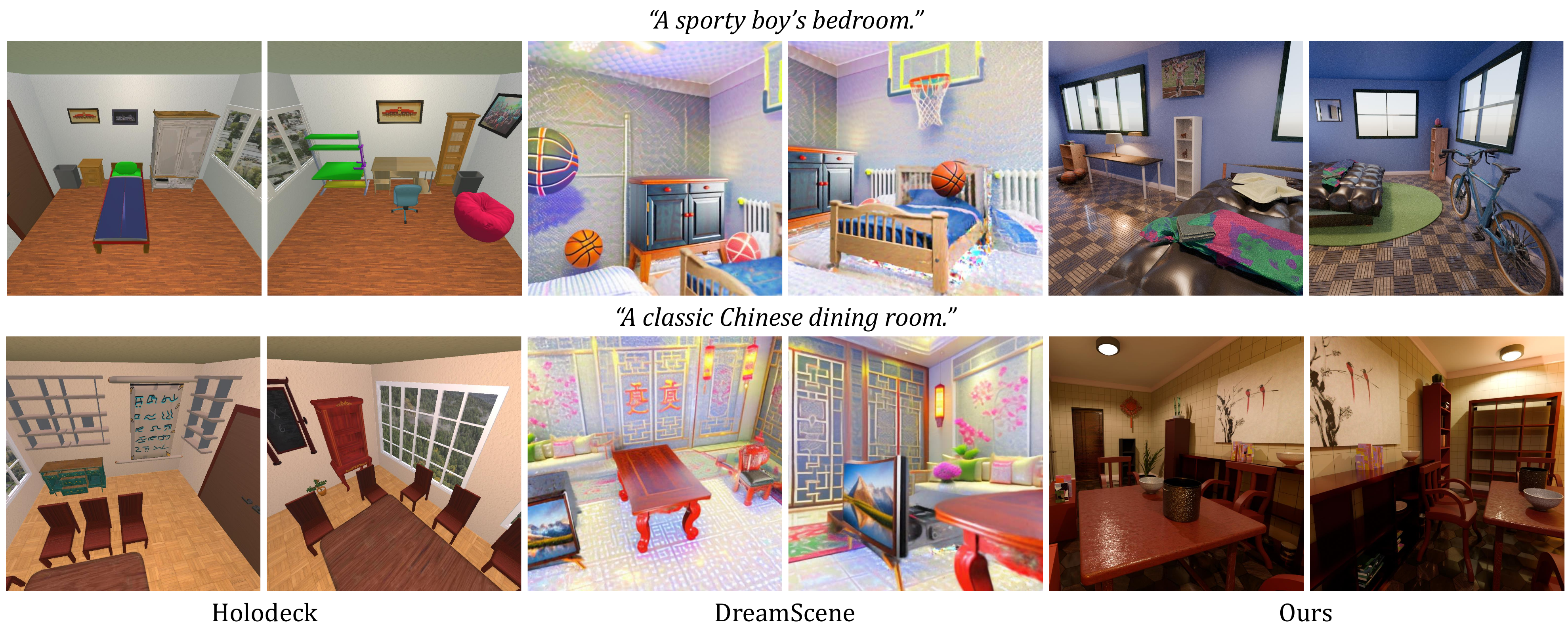}
\end{center}
\caption{\textbf{Qualitative comparison of 3D scene generation. } Compared with \cite{yang2024holodeck} and \cite{li2025dreamscene}, our method produces more realistic and visually consistent scenes, with accurate object placement and better adherence to the input text, demonstrating superior quality in both aesthetics and functionality.
}
\label{fig:comparison}
\end{figure*}

\subsection{Comparison}
As shown in \Cref{fig:comparison}, WorldCraft produces high-quality indoor 3D scenes. Compared to baseline approaches, our method produces more realistic appearance and geometry. Our approach generates layouts that are reasonable in terms of functionality, while the baselines may violate common sense, such as placing a basketball hoop over the bed. Moreover, our approach generates scenes with a style that is consistent with the input text, while automatically customizing objects, such as their material, texture, and shape. It also appropriately invokes a deep 3D generator to insert objects like a bicycle for ``a sporty boy's bedroom'' or a Chinese knot for ``a classic Chinese dining room,'' accurately reflecting the required styles. According to \Cref{tab:comparison}, our approach achieves the highest score, further validating its strength in consistency, aesthetics, and functionality. In addition, compared with diffusion-based baselines, our method is more efficient with a shorter runtime.

\begin{figure*}[t]
\begin{center}
    \includegraphics[width=0.8\linewidth]{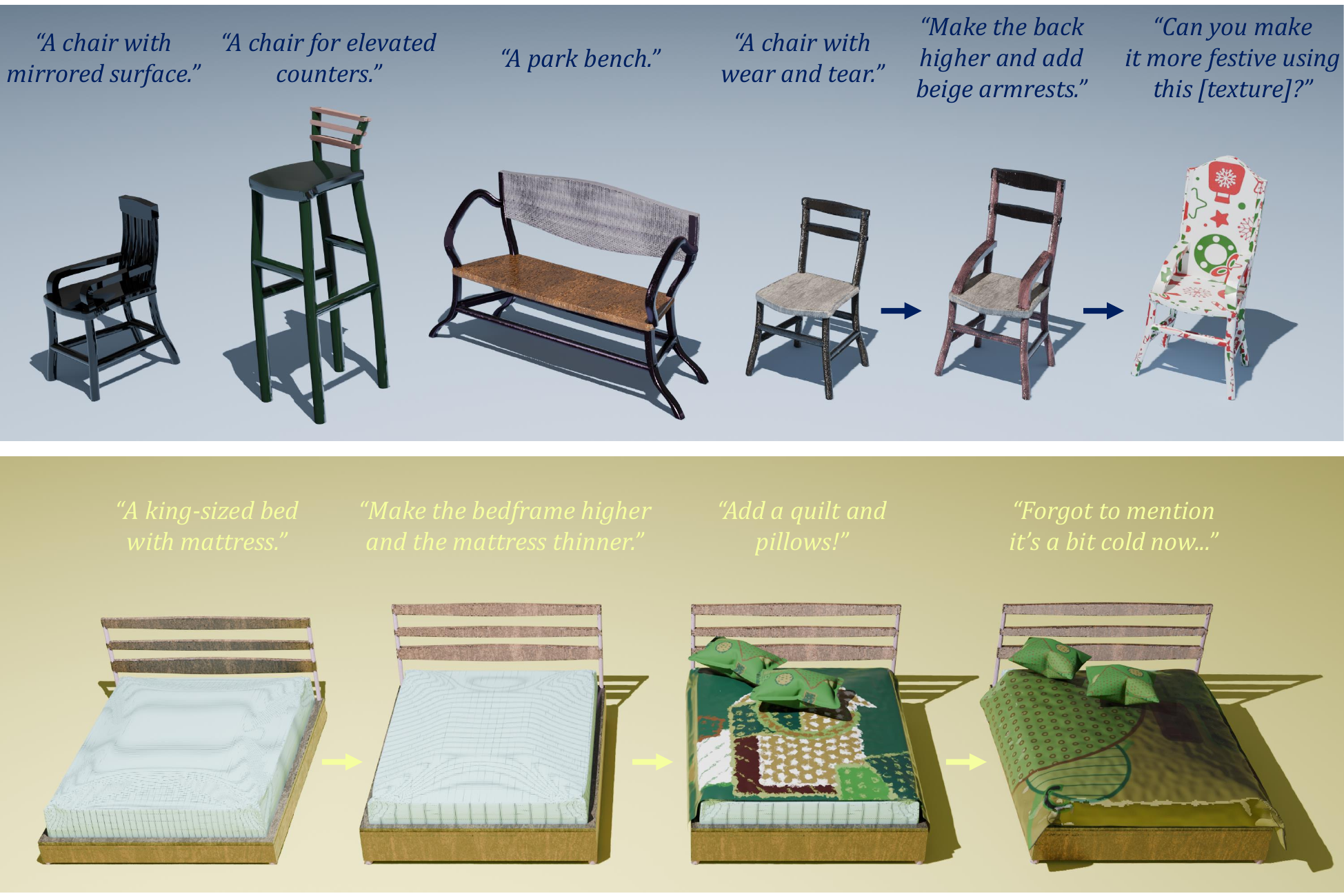}
\end{center}
\caption{\textbf{Qualitative results of ForgeIt.}
Our results demonstrate ForgeIt's ability to control object geometry and appearance. The user can refine and edit the results via multi-turn conversations and provide supplementary inputs, such as textures, to align with design intentions.
}
\label{fig:forgeit_result1}
\end{figure*}

\begin{figure*}[t]
\begin{center}
    \includegraphics[width=\linewidth]{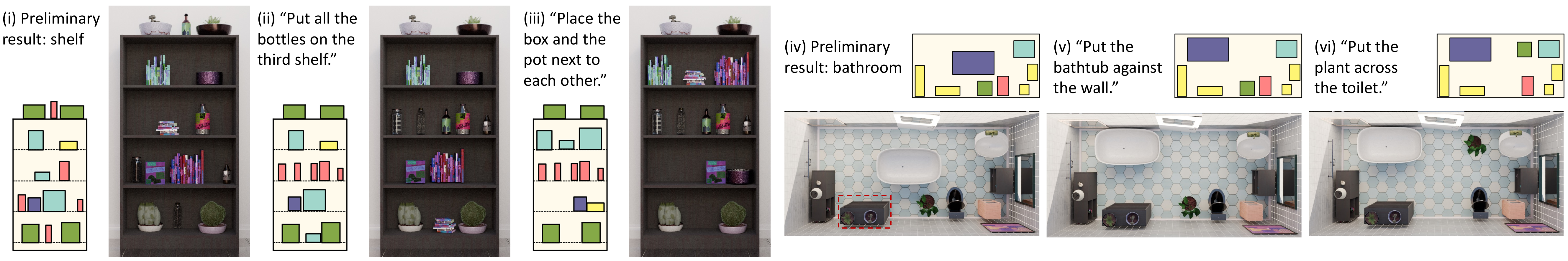}
\end{center}
\caption{\textbf{Qualitative results of ArrangeIt.} The agent extracts hierarchical relationships between objects and decomposes the task into several sub-problems to achieve a layout arrangement with good functionality and consistency with user requirements. The red box corresponds to the shelf in the first row.
}
\label{fig:arrangeit_result}
\end{figure*}

\noindent\textbf{Baselines.}
We compare our method to recent baseline methods for 3D scene generation: the LLM-based Holodeck~\cite{yang2024holodeck} and the diffusion-based DreamScene~\cite{li2025dreamscene}. Additionally, for the evaluation of our ArrangeIt module, we compare it with LayoutGPT~\cite{feng2024layoutgpt}, an LLM-based approach for generating layouts.

\subsection{Evaluation}
\noindent \textbf{ForgeIt.} In \Cref{fig:forgeit_result1}, we present qualitative results of the \textit{ForgeIt} module, demonstrating its ability to control object geometry and appearance through natural language. Users can engage in multi-turn conversations to progressively refine the generated results and provide supplementary inputs, such as textures, to better align with specific design intentions.
We also present quantitative validation of the dynamic manual construction procedure in ForgeIt. Specifically, we experiment with two variants: one without manual input (zero-shot generation) and one with a static user-coded manual. As shown in \Cref{tab:manual_procedures}, the ForgeIt module with dynamic manual reconstruction achieves the best performance in terms of consistency and aesthetics.

\begin{table}[t]
\centering
\caption{\textbf{Quantitative evaluation of the dynamic manual construction of ForgeIt.}
Our dynamic manual coaches the agent to master procedural generation, achieving the highest consistency and aesthetics score.}
\label{tab:manual_procedures}
\resizebox{\linewidth}{!}{%
\begin{tabular}{@{}ccccccc@{}}
\toprule
\multirow{2}{*}{{Manual Construction}} & \multicolumn{3}{c}{{Consistency}} & \multicolumn{2}{c}{{Aesthetics}} \\
\cmidrule(r){2-4} \cmidrule(r){5-6}
 & CLIP & User & GPT-4 & User & GPT-4 \\
\midrule
\xmark      & 0.271 & 4.59 & 6.50 & 5.60 & 7.50 \\
Static         & 0.273 & 4.31 & 6.00 & 5.57 & 7.50 \\
Dynamic        & \textbf{0.378} & \textbf{6.29} & \textbf{8.00} & \textbf{7.01} & \textbf{8.00} \\
\bottomrule
\end{tabular}}
\end{table}

\begin{table}[t]
\centering
\caption{\textbf{Quantitative evaluation of ArrangeIt.} ArrangeIt with hierarchical modeling generates a layout with the highest consistency and functionality scores.}

\label{tab:arrangeit}
\resizebox{\linewidth}{!}{%
\begin{tabular}{@{}lccccc@{}}
\toprule
\multirow{2}{*}{Layout Module} & \multicolumn{3}{c}{\textbf{Consistency}} & \multicolumn{2}{c}{\textbf{Functionality}} \\
\cmidrule(r){2-4} \cmidrule(r){5-6}
               &CLIP & User & GPT-4 & User & GPT-4 \\
\midrule
LayoutGPT            & 0.272    & 4.90 & 7.00 & 5.41 & 7.50 \\
ArrangeIt (w/o hierarchy)   & 0.340   &3.81 & 6.00 & 5.67 & 7.00 \\
ArrangeIt   & \textbf{0.361}   & \textbf{8.58} & \textbf{7.50} & \textbf{7.63} & \textbf{8.50} \\
\bottomrule
\end{tabular}}
\end{table}

\noindent \textbf{ArrangeIt.} In \Cref{fig:arrangeit_result}, we present qualitative results of the \textit{ArrangeIt} module, where the agent extracts hierarchical relationships between objects and decomposes the task into several sub-problems. For example, it separates the arrangement of smaller objects on a shelf from the placement of larger objects in a bathroom. For each sub-task, the user retains control, enabling layout adjustments at various levels. This capability results in a layout with good consistency to user requirements and functionality. This is further illustrated by the quantitative evaluation in \Cref{tab:arrangeit}, where we perform an ablation study on the layout module using LayoutGPT or ArrangeIt without hierarchical modeling.

\section{Conclusion}
 This work introduces WorldCraft, an LLM agent that utilizes procedural generation to create customizable indoor and outdoor scenes populated with various objects. With WorldCraft, users can interact using natural language to control individual object attributes and the overall scene layout. We propose ForgeIt, which develops an ever-growing manual through auto-verification to facilitate precise customization of individual objects. We also introduce ArrangeIt, which formulates hierarchical optimization problems to determine layouts that consider both ergonomic and aesthetic aspects. To complete our pipeline,  a trajectory control module is designed that enables users to animate the scene and operate the camera through natural language interactions. Our agent's 3D visual programming capabilities are compatible with off-the-shelf deep 3D generators for enhancing scene assets.
Our experiments demonstrate the versatility of WorldCraft in customizing complex 3D scenes and assisting non-professionals in realizing their creative visions.

{
\bibliographystyle{ACM-Reference-Format}

\def\CVPR{IEEE/CVF Conference on Computer Vision and Pattern Recognition (CVPR)}\def\ECCV{ European Conference on Computer Vision (ECCV)}\def\ICCV{IEEE/CVF International Conference on Computer Vision (ICCV)}\def\NIPS{Advances in Neural Information Processing Systems (NeurIPS)}\def\ICML{International Conference on Machine Learning (ICML)}\def\ICLR{International Conference on Learning Representations (ICLR)}\def\WACV{IEEE/CVF Winter Conference on Applications of Computer Vision (WACV)}\def\CVPRW{IEEE/CVF Conference on Computer Vision and Pattern Recognition (CVPR) Workshops}\def\ICCVW{IEEE/CVF International Conference on Computer Vision (ICCV) Workshops}\def\ICRA{IEEE International Conference on Robotics and Automation (ICRA)}\def\TOG{ACM Transactions on Graphics (TOG)}\def\PAMI{IEEE Transactions on Pattern Analysis and Machine Intelligence (PAMI)}\def\TIP{IEEE Transactions on Image Processing (TIP)}\def\IJCV{International Journal of Computer Vision (IJCV)}\def\SIGGRAPH{ACM Transactions on Graphics
  (SIGGRAPH)}\def\SIGGRAPHASIA{ACM Transactions on Graphics (SIGGRAPH Asia)}\def\TOG{ACM Transactions on Graphics (TOG)}\def\threedv{International Conference on 3D Vision (3DV)}\def\TVCG{IEEE Transactions on Visualization and Computer Graphics (TVCG)}\def\PMLR{Proceedings of Machine Learning Research (PMLR)}

}

\end{document}